\documentclass[conference,anonymous]{IEEEtran}
\IEEEoverridecommandlockouts
\usepackage{cite}
\usepackage{amsmath,amssymb,amsfonts}
\usepackage{algorithmic}
\usepackage{graphicx}
\usepackage{hyperref}
\usepackage{textcomp}
\usepackage{xcolor}
\usepackage{tikz}
\usetikzlibrary{arrows.meta}
\usepackage{subcaption}
\def\BibTeX{{\rm B\kern-.05em{\sc i\kern-.025em b}\kern-.08em
    T\kern-.1667em\lower.7ex\hbox{E}\kern-.125emX}}
\begin{document}

\title{Multifunctionality in a Connectome-Based Reservoir Computer\\

}
% \author{\IEEEauthorblockN{Anonymous Authors}
% \IEEEauthorblockA{\textit{}\\
% %\vspace{1.4cm}\\
% \vspace{-.7cm}\\
% }
% }

\author{
\IEEEauthorblockN{Jacob Morra}
\IEEEauthorblockA{\textit{Western University}\\
London, ON Canada \\
\texttt{\href{mailto:jmorra6@uwo.ca}{jmorra6@uwo.ca}}}
\and
\IEEEauthorblockN{Andrew Flynn}
\IEEEauthorblockA{\textit{University College Cork}\\
Cork, Ireland \\
\texttt{\href{mailto:andrew$\_$flynn@umail.ucc.ie}{andrew$\_$flynn@umail.ucc.ie}}}
\and
\IEEEauthorblockN{Andreas Amann}
\IEEEauthorblockA{\textit{University College Cork}\\
Cork, Ireland \\
\texttt{\href{mailto:a.amann@ucc.ie}{a.amann@ucc.ie}}}
\and
\IEEEauthorblockN{}
\IEEEauthorblockA{\textit{} \\
\textit{}\\
 \\
\texttt{}}
\and
\IEEEauthorblockN{Mark Daley}
\IEEEauthorblockA{\textit{Western University}\\
London, ON Canada \\
\texttt{\href{mailto:mdaley2@uwo.ca}{mdaley2@uwo.ca}}}
}

\maketitle
\begin{abstract}
Multifunctionality describes the capacity for a neural network to perform multiple mutually exclusive tasks without altering its network connections; and is an emerging area of interest in the reservoir computing machine learning paradigm. Multifunctionality has been observed in the brains of humans and other animals: particularly, in the lateral horn of the fruit fly. In this work, we transplant the connectome of the fruit fly lateral horn to a reservoir computer (RC), and investigate the extent to which this `fruit fly RC' (FFRC) exhibits multifunctionality using the `seeing double' problem as a benchmark test. We furthermore explore the dynamics of how this FFRC achieves multifunctionality while varying the network's spectral radius. Compared to the widely-used Erd\"{o}s–Renyi Reservoir Computer (ERRC), we report that the FFRC exhibits a greater capacity for multifunctionality; is multifunctional across a broader hyperparameter range; and solves the seeing double problem far beyond the previously observed spectral radius limit, wherein the ERRC's dynamics become chaotic.
\end{abstract}


\section{Introduction}

In the pursuit of developing artificially intelligent systems, there is much to be gained from integrating further physiological features of biological neural networks (BNNs) into machine learning (ML) environments. Inspired by the ability of certain BNNs to exhibit `multifunctionality', in this paper we investigate whether a corresponding connectome maintains multifunctionality when transplanted to the reservoir computing ML paradigm.

\textit{Multifunctionality} describes the ability of a neural network to perform multiple tasks without changing any network connections \cite{briggman08multifunctional}. This neurological phenomenon was first translated from biological to artificial neural networks (ANNs) using a `reservoir computer' (RC) in \cite{Flynn2021_mfrc_first}. An RC is a \textit{dynamical system} which can be realised as an ANN. What distinguishes the RC amongst other ML approaches is that network weights are trained at the \textit{readout layer only} in order to solve a given task. Multifunctional RCs, in particular, have the capacity to reconstruct a coexistence of chaotic attractors from: a multistable system \cite{Flynn2021_mfrc_first}; two different systems \cite{flynn2022_MFLimits}; and multiple copies of a chaotic attractor from the same system \cite{Flynn2021_symmetry}; all without re-training the weights or re-tuning hyperparameters.

In the life sciences, multifunctionality is well-documented. For example, in the land snail, the subesophageal ganglion complex has been shown to act as a controller for maintaining homeostasis of respiratory, genital, and cardiorenal functions \cite{Rozsa1987-fo}; in the turtle spinal cord, multifunctionality is present in the interneurons, which contribute to the forward swimming, flexion, and scratching reflexes \cite{Berkowitz2010-lk} -- analogous interneurons are also present in the mammalian central nervous system \cite{Jankowska2001-ql}. In the fruit fly brain, there is evidence supporting multifunctionality in multiple regions of interest (ROIs): in the wing, for example, where neurons of the ventral nerve cord are multifunctional for both song and flight \cite{OSullivan2018-ky}; and in the lateral horn, where sleep and odour aversion coexist \cite{Hsu2021-kq}, and where olfactory and visual stimuli are mediated simultaneously \cite{Gupta2012-ow}.

In this paper, we investigate whether a fruit fly connectome-based RC (FFRC) exhibits multifunctionality as observed in its biological counterpart neural network; specifically, by capturing the \textit{approximate} -- see Sec.\,\ref{ssec:FFRCderivation} -- network topology of the lateral horn ROI from the hemibrain dataset \cite{Scheffer2020-uc} and applying its RC analogue (as in \cite{Morra2022-ml, Morra2022-nz}) to the `seeing double' problem. The \textit{seeing double} benchmark test for exploring the limits of multifunctionality was first introduced in \cite{flynn2022_MFLimits} and further explored in \cite{flynn2023seeingdouble}; in this test, the RC is tasked with having to reconstruct a coexistence of two circular orbits ($\mathcal{C}_{A}$ and $\mathcal{C}_{B}$) which rotate in opposing directions (see Sec.\,\ref{ssec:seeingdouble}).

The rest of the paper is outlined as follows: In Sec.\,\ref{sec:Background} we introduce the particular RC formulation of interest, outline how it is trained to become multifunctional, and describe the specifics of the seeing double problem. Thereafter in Sec.\,\ref{sec:Methods}, we describe each of the RC network topologies used, detail how they are constructed, and also propose four experiments which aim to shed some light on the differences between how each achieves multifunctionality (across varying hyperparameters of interest). Furthermore in Sec.\,\ref{sec:results} we discuss the results of all experiments. Finally, in Sec.\,\ref{sec:conclusion}, we summarize the major experimental findings, describe the project limitations and caveats, and highlight future directions for investigation.

\section{Background}\label{sec:Background}

\subsection{Reservoir Computing}\label{ssec:RCbackground}
We use the continuous-time RC formulation devised in \cite{LuHuntOtt18RC} which was shown in \cite{Flynn2021_mfrc_first,Flynn2021_symmetry,flynn2022_MFLimits,flynn2023seeingdouble} to be capable of achieving multifunctionality and is expressed in the following equation:
\begin{align}
    \dot{\boldsymbol{r}}(t) = \gamma \left[ - \boldsymbol{r}(t) + \tanh{\left( \,\, \textbf{M} \, \boldsymbol{r}(t) + \sigma \textbf{W}_{in} \, \boldsymbol{u}(t) \,\, \right)} \right]. \label{eq:ListenRes}
\end{align}
Here $\boldsymbol{r}(t) \in \mathbb{R}^{N}$ is the state of the RC at a given time $t$ and $N$ is the number of neurons in the network. $\gamma$ is the decay-rate parameter. $\textbf{M} \in \mathbb{R}^{N \times N}$ is the adjacency matrix describing the internal layer of the RC. $\sigma$ is the input strength parameter and $\textbf{W}_{in} \in \mathbb{R}^{N \times D}$ is the input matrix (constructed as in \cite{LuHuntOtt18RC}), when multiplied together this represents the weight given to the $D$-dimensional input, $\boldsymbol{u}(t) \in \mathbb{R}^{D}$, as it is projected into the RC. This input is taken from the particular attractor or time series that one would like to either reconstruct or make future predictions of. Solutions of Eq.\,\eqref{eq:ListenRes} are computed using the 4$^{th}$ order Runge-Kutta method with time step $\tau = 0.01$.

In our numerical experiments we consider two different variations of $\textbf{M}$. The first $\textbf{M}$ is constructed with an Erd\"{o}s-Renyi topology where each of the non-zero elements are then replaced with a random number between $-1$ and $1$; the matrix is subsequently scaled to a specific spectral radius, $\rho$. The second $\textbf{M}$ is constructed from the right lateral horn ROI from the hemibrain connectome \cite{Scheffer2020-uc}: further details on this construction are outlined in Sec.\,\ref{ssec:FFRCderivation}. The spectral radius, $\rho$, for each $\textbf{M}$ is a key parameter involved in the training: in particular, $\rho$ is associated with the RC's memory as it is used to tune the weight the RC places on its own internal dynamics.

To train the RC in Eq.\,\eqref{eq:ListenRes}, the system is first driven by the input $\boldsymbol{u}(t)$ from $t=0$ to time $t=t_{listen}$ in order to remove any dependency which $\boldsymbol{r}(t)$ has on its initial condition $\boldsymbol{r}(0)=\left( 0, 0, \ldots, 0 \right)^{T} = \mathbf{0}^{T}$. The training data is then generated by driving the RC with $\boldsymbol{u}(t)$ from $t=t_{listen}$ to $t=t_{train}$.

A suitable readout layer needs to be calculated in order to train the RC and replace the training input signal, $\boldsymbol{u}(t)$, in Eq.\,\eqref{eq:ListenRes} with a post-processing function, $\hat{\psi}\left( \cdot \right)$. If the training is successful then we say that,
\begin{align}
    \hat{\psi}\left(\boldsymbol{r}(t)\right) = \hat{\boldsymbol{u}}(t) \approx \boldsymbol{u}(t), \,\,\,\text{for}\,\, t > t_{train},
\end{align}
where $\hat{\boldsymbol{u}}(t)$ denotes the predicted time-series. This layer `closes the loop' of the nonautonomous system in Eq.\,\eqref{eq:ListenRes} and provides a map from the $N$-dimensional state space of the RC, $\mathbb{S}$, to the $D$-dimensional `prediction state space', $\mathbb{P}$.

In this work $\hat{\psi}\left(\boldsymbol{r}(t)\right) = \textbf{W}_{out} \boldsymbol{q}( \boldsymbol{r}(t) )$ -- where $\textbf{W}_{out}$ is the readout matrix -- and we use $\boldsymbol{q}( \boldsymbol{r}(t) )$ to break the symmetry in Eq.\,\eqref{eq:ListenRes} using the `squaring technique' described by
\begin{align}
    \boldsymbol{q}(\boldsymbol{r}(t))=\left(\begin{array}{cc}
\boldsymbol{r}(t), & \boldsymbol{r}^{2}(t)
\end{array}\right)^{T}.\label{eq:q_square}
\end{align}

We calculate $\textbf{W}_{out}$ using the ridge regression approach,
\begin{align}
    \textbf{W}_{out} = \textbf{Y} \textbf{X}^{T} \left( \textbf{X} \textbf{X}^{T} + \beta \, \textbf{I} \right)^{-1},\label{eq:WoutRegression}
\end{align}
where 
\begin{align}
\textbf{X} \hspace{-0.1cm} = \hspace{-0.1cm} \left[ \begin{array}{@{}c@{\hspace{1mm}}c@{\hspace{1mm}}c@{\hspace{1mm}}c@{}}
\left(\begin{array}{@{}c@{}}
\boldsymbol{r}(t_{listen})\\
\boldsymbol{r}^{2}(t_{listen})
\end{array}\right)
& \hspace{-0.14cm} \left(\begin{array}{@{}c@{}}
\boldsymbol{r}(t_{listen} + \tau)\\
\boldsymbol{r}^{2}(t_{listen} + \tau)
\end{array}\right) & \hspace{-0.13cm} \cdots 
& \hspace{-0.13cm} \left(\begin{array}{@{}c@{}}
\boldsymbol{r}(t_{train})\\
\boldsymbol{r}^{2}(t_{train})
\end{array}\right)
\end{array} \right]\label{eq:Xmat}
\end{align}
is the RC reservoir's response to the input data, which is represented as
\begin{align}
    \textbf{Y} = \left[ \begin{array}{cccc}
        \boldsymbol{u}(t_{listen}) & \boldsymbol{u}(t_{listen} + \tau) & \cdots & \boldsymbol{u}(t_{train})
    \end{array} \right].\label{eq:Ymat}
\end{align}
In Eq. \ref{eq:WoutRegression}$, \beta$ is the regularization parameter which is used to help prevent overfitting. $\textbf{I}$ is the identity matrix.

We write the `predicting RC' as the following autonomous dynamical system:
\begin{align}
\dot{\hat{\boldsymbol{r}}}(t) \hspace{-0.05cm} = \hspace{-0.05cm} \gamma \Big[ \hspace{-0.05cm} - \hat{\boldsymbol{r}}(t) \hspace{-0.05cm} + \hspace{-0.05cm} \tanh \hspace{-0.05cm} \Big( \textbf{M} \, &\hat{\boldsymbol{r}}(t) \hspace{-0.05cm} + \hspace{-0.05cm} \sigma \textbf{W}_{in} \textbf{W}_{out}^{(1)} \boldsymbol{q}(\hat{\boldsymbol{r}}(t)) \Big) \Big], \label{eq:PredRes}
\end{align}
where $\hat{\boldsymbol{r}}$ denotes the state of the predicting RC at time $t$ and $\hat{\boldsymbol{r}}(0) = \boldsymbol{r}(t_{train})$.

For the case of multifunctionality, Eq.\,\eqref{eq:ListenRes} is driven by two different input signals, $\boldsymbol{u}_{1}$ and $\boldsymbol{u}_{2}$, that describe trajectories on two attractors $\mathcal{A}_{1}$ and $\mathcal{A}_{2}$. Following the above steps, this produces two corresponding RC response data matrices, $\textbf{X}_{1}$ and $\textbf{X}_{2}$, and in accordance with the input data matrices, $\textbf{Y}_{1}$ and $\textbf{Y}_{2}$. These $\textbf{X}_{1}$ and $\textbf{X}_{2}$, and $\textbf{Y}_{1}$ and $\textbf{Y}_{2}$ are `blended' together according to the \textit{blending technique} featured in \cite{Flynn2021_mfrc_first}. The resulting blended matrices are used to solve for $\textbf{W}_{out}$ according to Eq.\,\eqref{eq:WoutRegression}. The predicting RC in this case is described as in Eq.\,\eqref{eq:PredRes}; if multifunctionality is achieved, once Eq.\,\eqref{eq:PredRes} is initialised with either $\hat{\boldsymbol{r}}_{1}(0)$ or $\hat{\boldsymbol{r}}_{2}(0)$ then the predicting RC will reconstruct the dynamics of either $\mathcal{A}_{1}$ or $\mathcal{A}_{2}$.

\subsection{The `Seeing Double' task}\label{ssec:seeingdouble}

\begin{figure}[t]
    \centering
    \begin{tikzpicture}[scale=1.7]
    \draw[orange] (0, 0) circle (1cm);
    \draw[-{Stealth[length=5pt]}, thick] (-1.1,0.27) arc (165:90:1.1cm);
    \node at (0, 0.6) {$\mathcal{C}_{B}$};
    \fill (0., 0) circle (2pt);
    \draw[blue]  (1.5, 0) circle (1cm);
    \draw[-{Stealth[length=5pt]}, thick] (0., -0.5) -- (0., -0.1);
    \node at (0, -0.65) {$(-x_{cen}, -y_{cen})$};
    \draw[-{Stealth[length=5pt]}, thick] (1.5, -1.1) arc (270:335:1.1cm);
    \node at (1.5, 0.6) {$\mathcal{C}_{A}$};
    \fill (1.5, 0) circle (2pt);
    \draw[-{Stealth[length=5pt]}, thick] (1.5, -0.5) -- (1.5, -0.1);
    \node at (1.5, -0.65) {$(x_{cen}, y_{cen})$};
    \draw[-{Stealth[length=5pt]}, thick] (1.5, 0.) -- (2.5, 0.);
    \node at (2., 0.1) {$s$};
    \end{tikzpicture}
    \caption{Illustration of the fundamentals of the seeing double problem.}
    \label{fig:seeingdouble}
\end{figure}
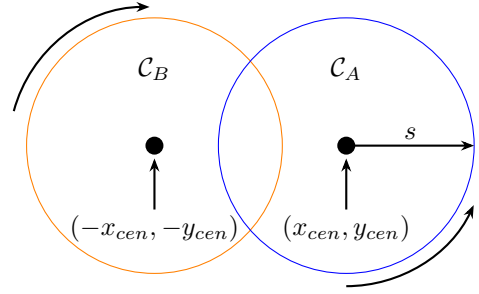

The \textit{seeing double} task was first introduced in \cite{flynn2022_MFLimits} as a benchmark task to compare how different RCs achieve multifunctionality. An illustration of the basic setup which is used in this numerical experiment is provided in Fig. \ref{fig:seeingdouble}.

For this task we consider training the RC in Eq.\,\eqref{eq:ListenRes} to reconstruct a coexistence of attractors which describe trajectories on two (partially or completely) overlapping circular orbits -- $\mathcal{C}_{A}$ and $\mathcal{C}_{B}$ -- that rotate in opposite directions (see Fig. \ref{fig:seeingdouble}). The input data to train the RC in Eq.\,\eqref{eq:ListenRes} is generated by
\begin{equation}\label{eqn:seeingdouble}
u(t) = \bigg(\begin{array}{c}
     x(t)  \\
     y(t)
\end{array}\bigg)
= \bigg(\begin{array}{c}
     s_x\,\cos(t) + x_{cen}  \\
     s_y\,\sin(t) + y_{cen}
\end{array}\bigg).
\end{equation}
In this paper, to train the RC in Eq.\,\eqref{eq:ListenRes} to become multifunctional, $s_{x}$ and $s_{y}$ are assigned as $s_{x}=s_{y}=5$ to create $\mathcal{C}_{A}$ and thus the training input signal $\boldsymbol{u}_{1}$; moreover, for $\mathcal{C}_{B}$ we use $s_{x}=-5$ and $s_{y}=5$ to produce the corresponding $\boldsymbol{u}_{2}$. In this case, the radius ($s$) of both $\mathcal{C}_{A}$ and $\mathcal{C}_{B}$ is equal to $5$.

Formally, the RC achieves multifunctionality in this instance if -- after training on the input time series from Eq.\,\eqref{eqn:seeingdouble} -- it reconstructs a coexistence of attractors in $\mathbb{S}$ such that the dynamics in $\mathbb{P}$ resembles trajectories on both $\mathcal{C}_{A}$ and $\mathcal{C}_{B}$. In practice we determine whether the `reconstructed attractors' in $\mathbb{P}$ -- that is, $\hat{\mathcal{C}}_{A}$ and $\hat{\mathcal{C}}_{B}$ -- follow the correct directional arcs and satisfy a `roundness' condition below $0.25$ -- determined empirically in \cite{flynn2022_MFLimits,flynn2023seeingdouble} and is defined as the difference in radii between the largest and smallest circular trajectories which inscribe a particular circular orbit. 

%{\jm{Could trim this paragraph}}

While the seeing double problem may at first seem as a relatively simple problem -- i.e. in comparison to reconstructing chaotic attractors -- it is the overlap between $\mathcal{C}_{A}$ and $\mathcal{C}_{B}$ that makes this task difficult for the RC to solve. For instance, when the RC approaches a junction where the circular trajectories intersect, it must use its \textit{memory} of the previous time steps in order to continue on its correct trajectory. %What is most remarkable is that even in the extreme scenario where the training data is completely overlapping the RC can still `see double' for a small range of $\rho$ values.

In \cite{flynn2022_MFLimits,flynn2023seeingdouble}, the critical role of $\rho$ is revealed: in particular, it is found that when $\mathcal{C}_{A}$ and $\mathcal{C}_{B}$ are completely overlapping -- i.e. when $x_{cen}=y_{cen}=0$ -- then multifunctionality is achieved for a small range of relatively large values of $\rho$. On the other hand, if $\rho$ is too large, then multifunctionality is lost. 

To provide a comparison, in this paper we confine our study to the this extreme scenario where $x_{cen}=y_{cen}=0$. %We show in Sec.\,\ref{sssec:Exp4results_bifplots} that this can lead towards the RC exhibiting chaos. 
%question: should we be sharing this result at this point, or save it for that particular section

\section{Methods}\label{sec:Methods}

\subsection{Model Pipeline}
\subsubsection{The Erd\"{o}s–Renyi Reservoir Computer (ERRC)}\label{ssec:ERRCderivation}
As mentioned in Sec.\,\ref{ssec:RCbackground}, we use a weighted Erd\"{o}s–Renyi topology with a sparsity of $0.05$. Weights are drawn randomly from $[-1,1]$ to construct an adjacency matrix $\textbf{M}_{ER}$ of size $N=500$. We vary the spectral radius, $\rho$, of the resulting $\textbf{M}_{ER}$ from 0 to 2.0 for Experiments 1-2 and 0 to 1.8 for Experiment 3 as outlined in Secs.\,\ref{sssec:Exp1_outline}-\ref{sssec:Exp3_outline}. We also explore the ERRC's dynamics at larger $\rho$ values (up to $\rho=2.2$) in Experiment 4.

\subsubsection{The Fruit Fly Reservoir Computer (FFRC)}\label{ssec:FFRCderivation}
The ``Fruit Fly RC'' (FFRC) is derived from the \textit{hemibrain} \cite{Scheffer2020-uc}: Using Neuprint \cite{noauthor_undated-qx}, we access the hemibrain API with a provided key, and run a set of Cypher queries on the Neo4j graph database to select all neurons in the right lateral horn ROI which are \textit{connected} to all others (i.e. in the same ROI). We define two neurons as being ``connected'' if they are synaptic partners and have a sufficiently-large number of shared synaptic sites; here we select a tolerance of 50 synapses in order to reduce the size and complexity of the model -- i.e. to make the training pipeline more efficient. After querying the hemibrain to collect neuron connection data, we construct a NetworkX graph which retains all synaptic partners and their \textit{weights} (the number of synaptic sites); finally, from this graph we construct our adjacency matrix $\textbf{M}_{FF}$ of size $N=426$. We interpolate the weight values into $[-1,1]$. Following a common approach (as in \cite{Gao2020-xm} and \cite{Proske2012-yt}), we also diagonalize the matrix. It is important to stress that $\textbf{M}_{FF}$ is unchanging: it has a \textit{fixed structure}, which is based off of the hemibrain connectome data, whereas the ERRC by contrast is randomly initialised.

\subsection{Experiments}\label{ssec:Exps_outline}

In each of the experiments listed below, we set $t_{listen}=6 T$ and $t_{train}=15 T$, where $T$ is the period of rotation on each orbit $\mathcal{C}$. We also assign $\sigma = 0.2$ and $\beta = 0.01$.

\subsubsection{Experiment 1 -- Multifunctionality trials}\label{sssec:Exp1_outline}
We conduct 50 sets of 100 trials; in each set, the ERRC and FFRC are both independently evaluated on the seeing double problem for previously-found optimal hyperparameter values ($\rho=1.4$, $\gamma=5.0$) -- see \cite{flynn2022_MFLimits}. $\textbf{W}_{in}$ is randomly initialised on each simulation for both the RC setups. While the FFRC, as previously mentioned, uses a fixed $\textbf{M}$ structure, the corresponding $\textbf{M}_{ER}$ for the ERRC is re-initialized for each trial.

In a given set of trials, we count all instances where multifunctionality is achieved according to the criteria in Sec.\,\ref{ssec:seeingdouble} and as outlined in \cite{flynn2022_MFLimits,flynn2023seeingdouble}. %We designate this count as representing a given model's \textit{performance}.

\subsubsection{Experiment 2 -- Varying $\rho$ and $\gamma$}\label{sssec:Exp2_outline}
As previously observed in \cite{flynn2022_MFLimits}, depending on the choice of both $\rho$ and $\gamma$, these parameters can have a profound impact on the multifunctionality of RCs on the seeing double problem. Here, as in \cite{flynn2022_MFLimits}, we aim to determine the regions in the $(\rho, \gamma)$-plane where multifunctionality is achieved for both the FFRC and ERRC. We conduct one set of 100 multifunctionality trials (as in Experiment 1) for each $\rho, \gamma$ combination for $\rho, \gamma$ $\in$ $[0, 2.0] \times [5, 95]$.

\subsubsection{Experiment 3 -- Comparing RC activations}\label{sssec:Exp3_outline} As the RC's predicted time series is a trained linear combination of RC activation states, it is reasonable to analyze the activity of each neuron in the cases of multifunctionality and non-multifunctionality, for both RC models.

Following this intuition, we compute the number of \textit{unique local maxima} -- on the interval from $t=t_{train}$ to $t=27T$ -- for each neuron ($\hat{r}_{(i)}$) in both RC setups. We then construct a \textit{heat map} (see Fig. \ref{fig:maxcounts}) which shows the count for each neuron versus $\rho$ for a multifunctional and a \textit{non-multifunctional} -- i.e. in cases where only one orbit is reconstructed or neither -- instance of the FFRC and ERRC models, respectively.

\subsubsection{Experiment 4 -- Exploring seeing double dynamics}\label{sssec:Exp4_outline}
Finally, we aim to shed some light on the differences between how the FF and ER RCs solve the seeing double problem, which follows the bifurcation analysis presented in \cite{flynn2023seeingdouble} -- here we explore the changes in the dynamics of the FF and ER RCs in $\mathbb{P}$ for changes in $\rho$. More specifically, we track the evolution of predictions in the $[\rho, x, y]$-space and illustrate how $\hat{\mathcal{C}}_{A}$ and $\hat{\mathcal{C}}_{B}$ come into existence in cases where both the FF and ER RCs achieve multifunctionality. The influence of `untrained attractors', attractors which exist in $\mathbb{P}$ but were not present during the training (like in \cite{flynn2023seeingdouble,Flynn2021_mfrc_first}) are also examined here. 

\section{Results and Discussion}\label{sec:results}

Note: we recommend the use of `Adobe Reader' or `Chrome PDF Viewer' to view the figures in this section.

\subsubsection{Experiment 1}
\begin{figure}[t!]
    \centering
    \includegraphics[width=.37\textwidth]{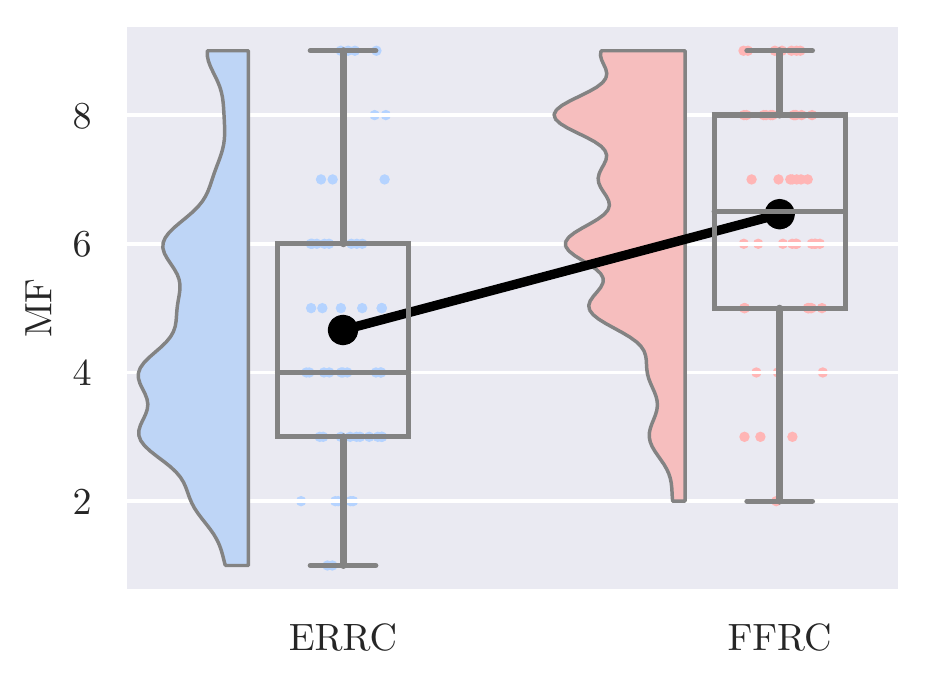}
    \caption{A raincloud plot illustrating FF versus ER RC multifunctionality (MF) across 50 sets of 100 trials on the seeing double problem.}
    \label{fig:combplot}
\end{figure}

Comparing the instances where multifunctionality (MF) occurs in the FF and ER RCs on the seeing double problem across 50 sets of 100 trials -- see Fig. \ref{fig:combplot}) -- we observe that the FFRC achieves an average multifunctionality count of 6.46 out of 100; conversely, the RC scores 4.66 out of 100 on average. The distribution of FFRC scores is neither positively nor negatively skewed; whereas the RC is negatively skewed. Differences observed between the distributions of multifunctionality instances are significant: here $p = 0.000035 < 0.05$ for the Wilcoxon rank-sum test.

\subsubsection{Experiment 2}

\begin{figure}[t!]
    \centering
    \includegraphics[width=.45\textwidth]{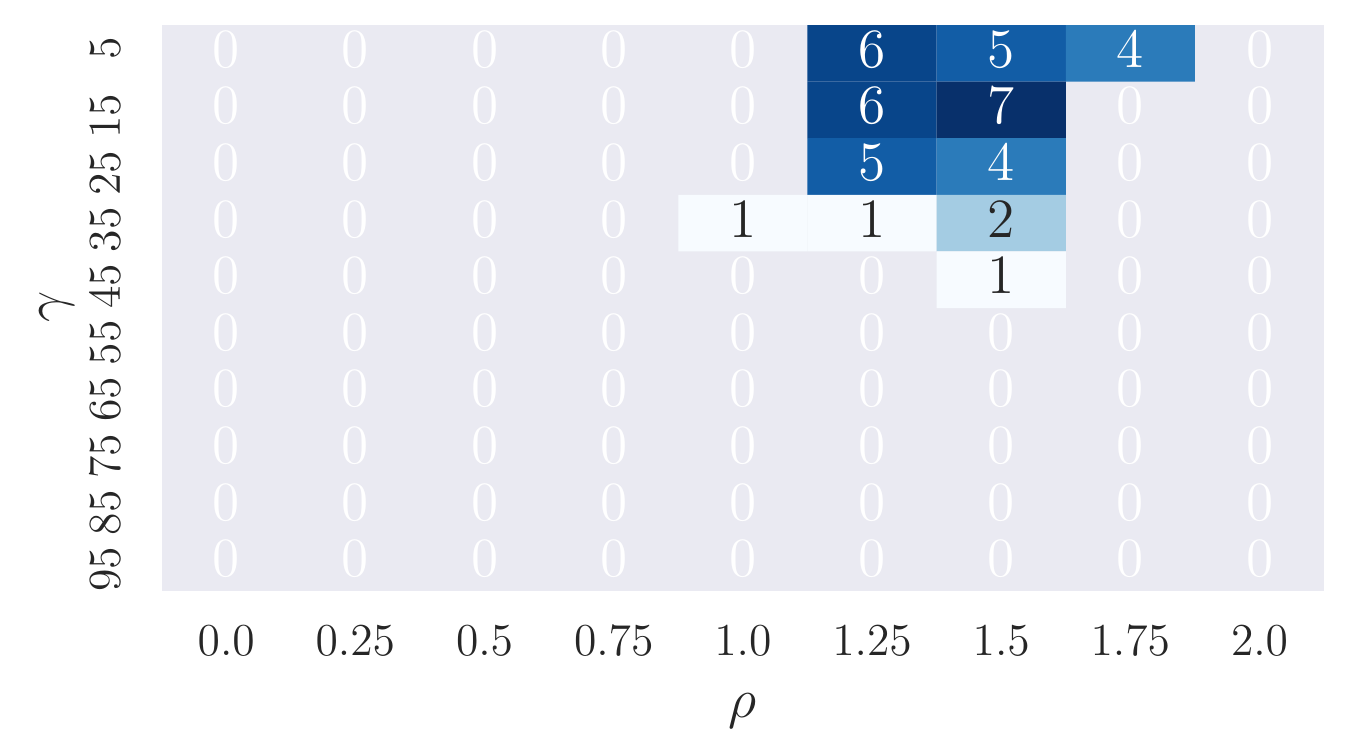}
    \caption{ERRC counts of multifunctionality in the $[\gamma, \rho]$-plane.}
    \label{fig:heatmapRand}
\end{figure}
\begin{figure}[t!]
    \centering
    \includegraphics[width=.45\textwidth]{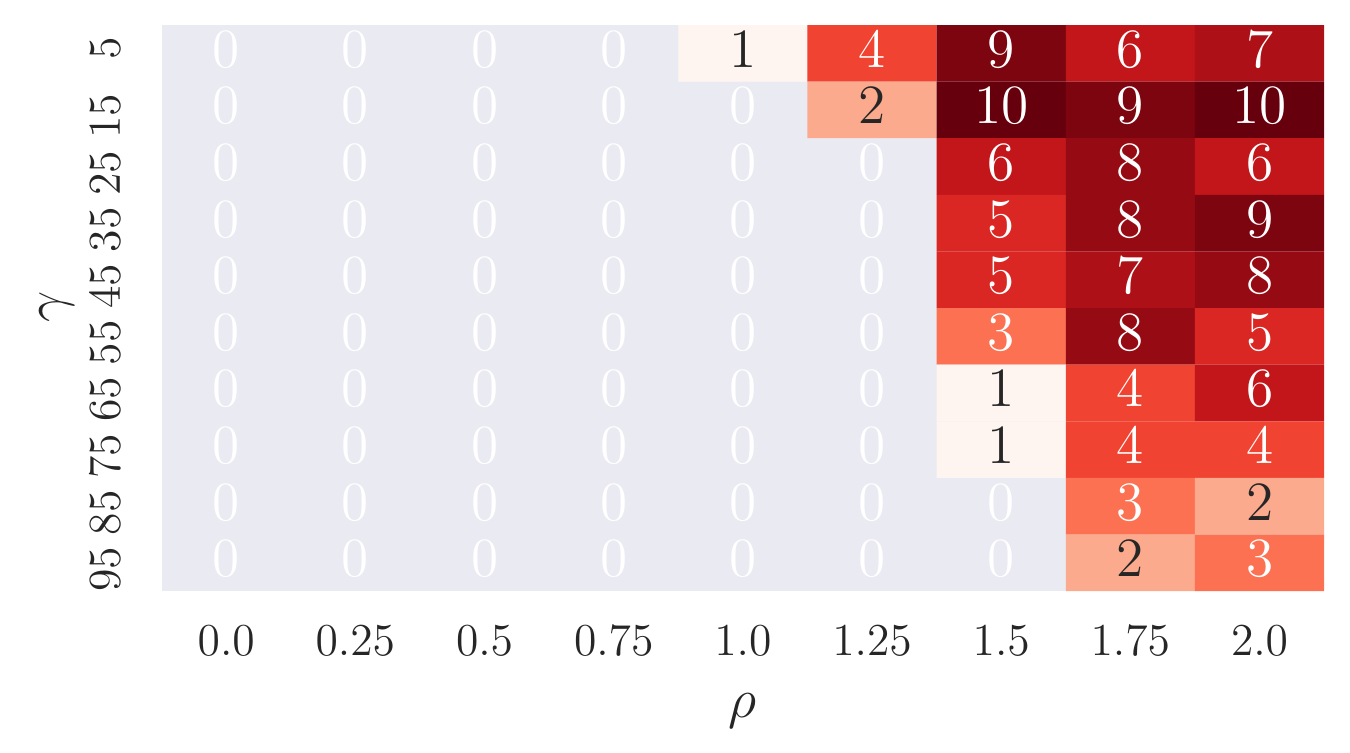}
    \caption{FFRC counts of multifunctionality in the $[\gamma, \rho]$-plane.}
    \label{fig:heatmapFFRC}
\end{figure}

In Fig. \ref{fig:heatmapRand} we provide multifunctionality counts (out of 100 trials, respectively) for the ERRC on the seeing double problem across the $[\gamma, \rho]$-plane of $[5,95] \times [0,2.0]$. We report an approximate window of multifunctionality in the $[\gamma$, $\rho]$ range of $[5,35] \times [1.25,1.5]$. Maximum multifunctionality occurs at $[\gamma, \rho]$=$[15,1.5]$ -- here we observe a count of 7 out of 100 trials. As previously observed in \cite{flynn2022_MFLimits}, multifunctional capacity falls as $\rho$ continues to increase. At $\rho=2.0$, for example, we observe no evidence of multifunctionality for any of the reported $\gamma$ values.

For the FFRC (see Fig. \ref{fig:heatmapFFRC}) we find the greatest occurrance of multifunctionality at $[\gamma, \rho] = [15,1.5]$ and also $[15,2.0]$, where the reported count is 10 instances out of 100 trials. The hyperparameter region where multifunctionality occurs is approximately $[\gamma, \rho]$ $\in$ $[5,75]$ $\times$ $[1.5,2.0]$, a much larger than the window of multifunctionality found for the ERRC. Moreover, the frequency of multifunctionality across all sets of 100 trials is greater overall. Importantly, the FFRC is also capable of exhibiting multifunctionality at large $\rho$ values (the ERRC is not); which is particularly interesting as this is where we observe -- i.e. for $\gamma=15$ -- \textit{maximum multifunctionality}.

\subsubsection{Experiment 3}
\begin{figure*}[ht]
\centering
\begin{subfigure}[h]{0.245\linewidth}
\centering
\includegraphics[width=0.945\linewidth]{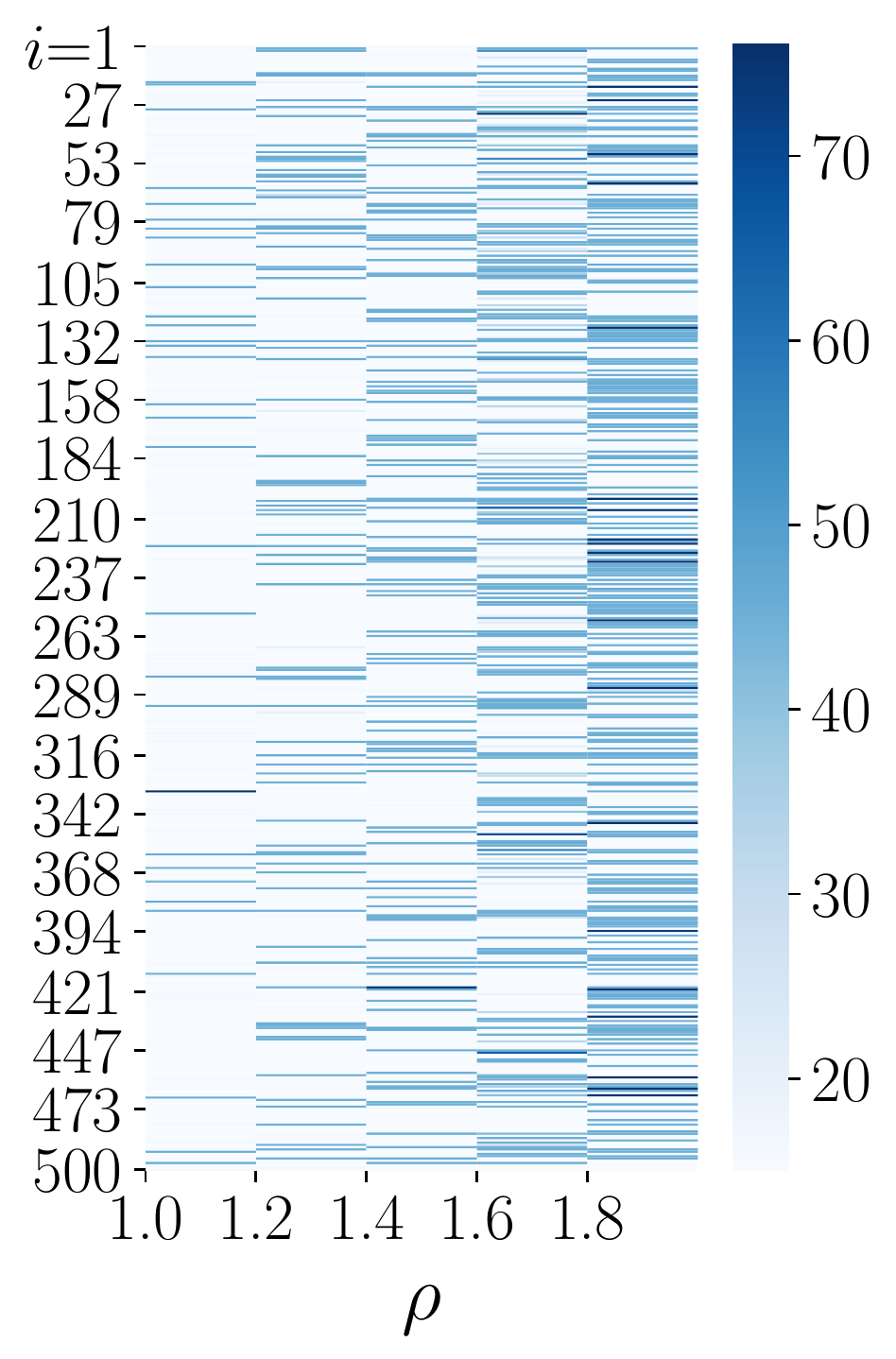} 
\caption{ERRC (MF)}
\label{ffrcMF}
\end{subfigure}
% \hspace{0.02\linewidth}
\begin{subfigure}[h]{0.245\linewidth}
\centering
\includegraphics[width=0.945\linewidth]{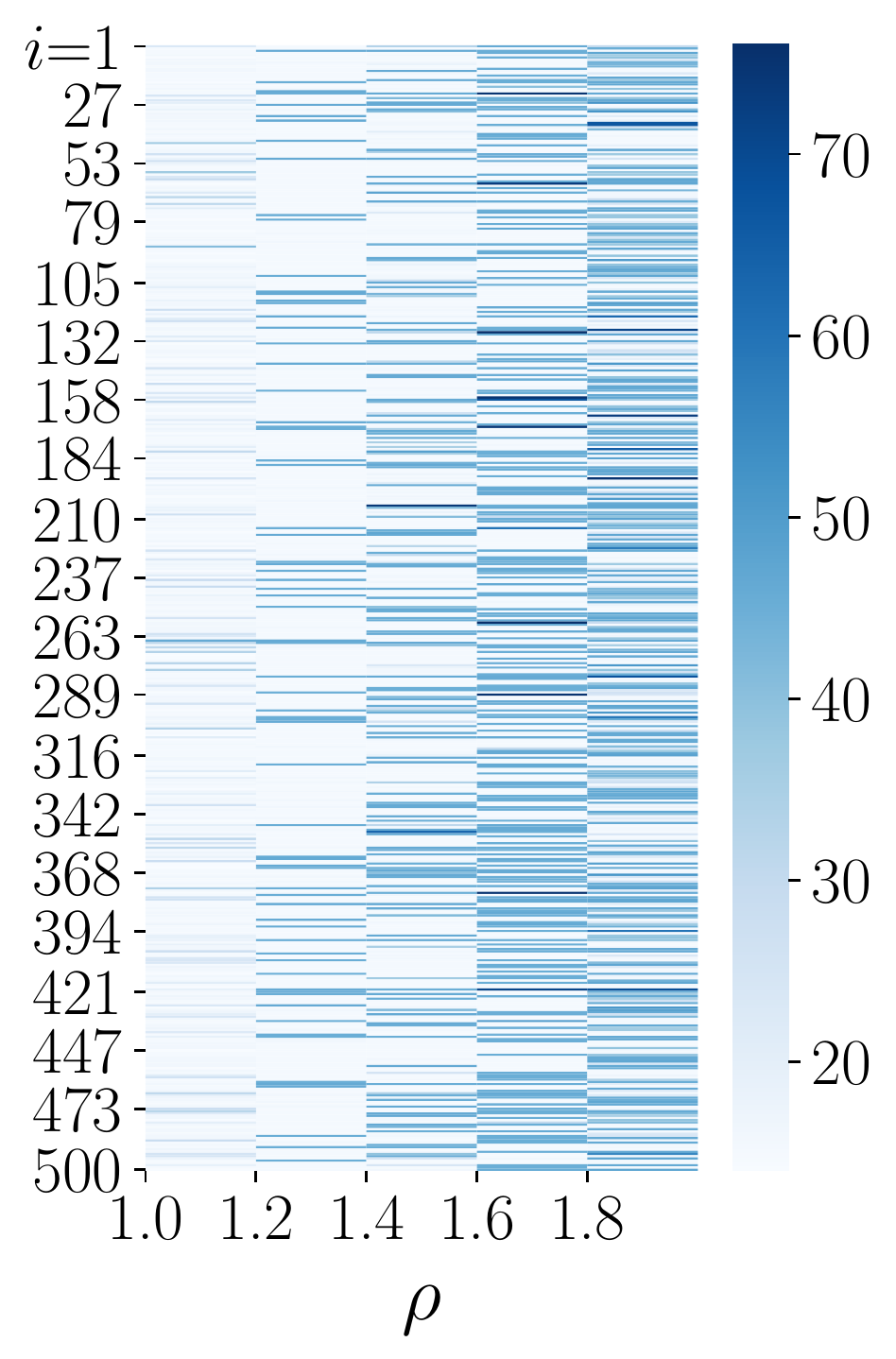} 
\caption{ERRC (non-MF)}
\label{ffrcnonMF}
\end{subfigure}
% \hspace{0.02\linewidth}
\begin{subfigure}[h]{0.245\linewidth}
\centering
\includegraphics[width=0.945\linewidth]{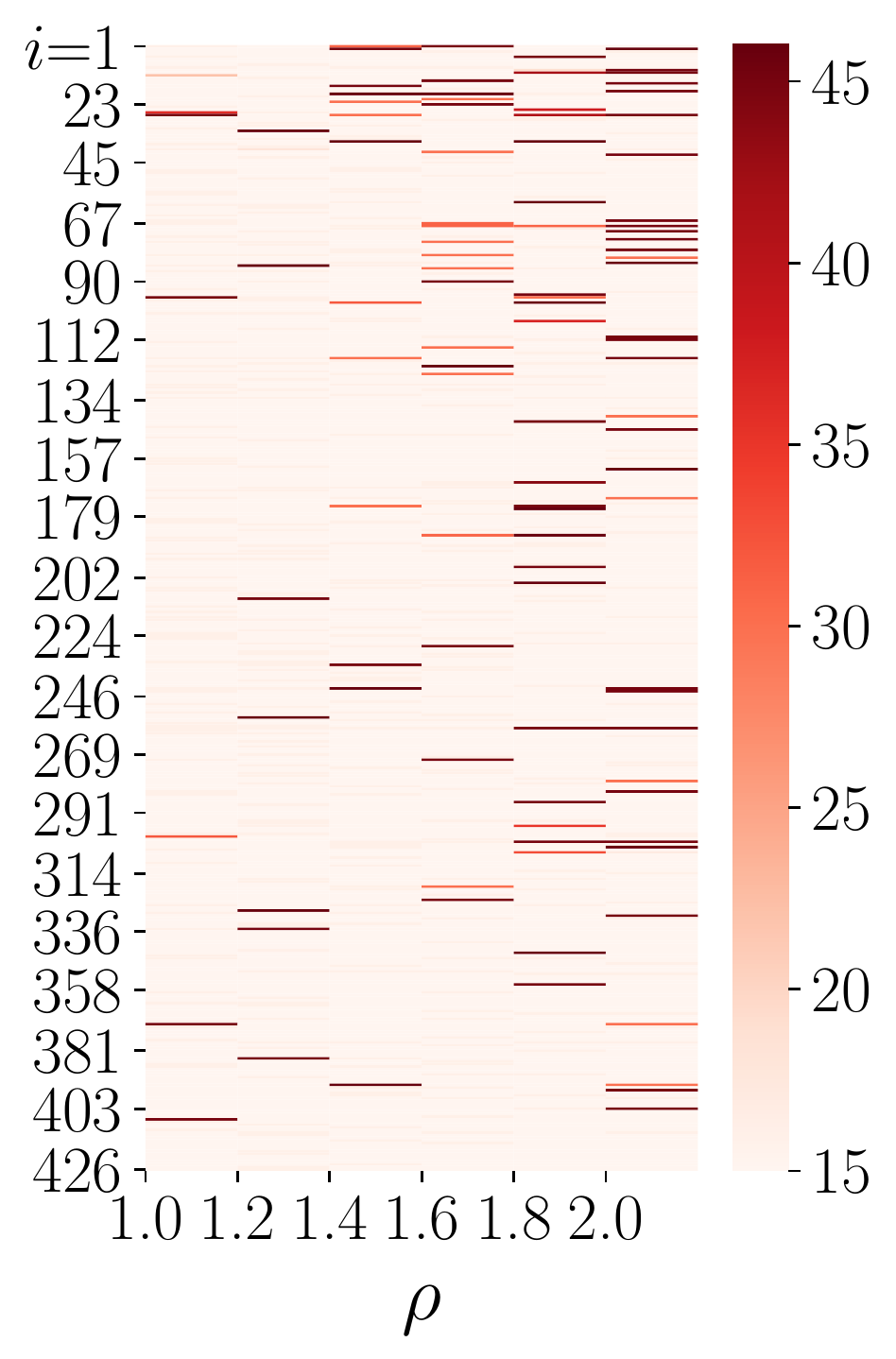} 
\caption{FFRC (MF)}
\label{randMF}
\end{subfigure}
% \hspace{0.02\linewidth}
\begin{subfigure}[h]{0.245\linewidth}
\centering
\includegraphics[width=0.945\linewidth]{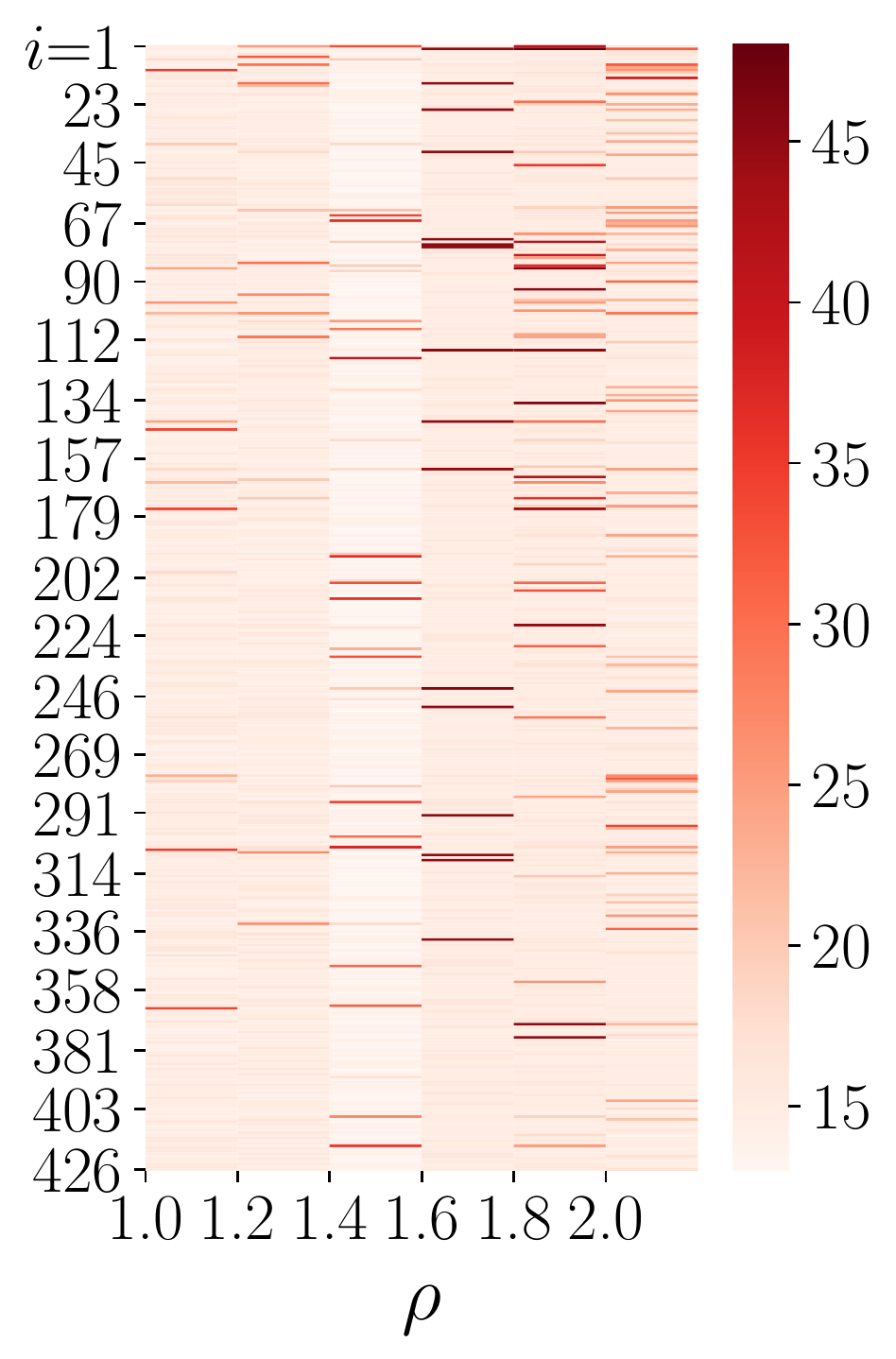} 
\caption{FFRC (non-MF)}
\label{randnonMF}
\end{subfigure}
\caption{Unique local maxima counts for each $i^{th}$ neuron vs. $\rho$ in a case where multifunctionality (MF) is and is not achieved for the FFRC in (a) and (b) and the ERRC in (c) and (d).}
\label{fig:maxcounts}
\end{figure*}

Looking at the activity profiles of all neurons $\hat{r}_{(i)}$ in the ERRC and FFRC (Fig. \ref{fig:maxcounts}), we first compare the population dynamics of RCs exhibiting multifunctionality (MF) and non-multifunctionality. For the FFRC activations, we observe that the number of unique local maxima is ``high'' (above 40) in more neurons during MF. We speculate -- as in \cite{Gallicchio2020-fl} -- that a higher proportion of reservoir activation neurons with many unique local maxima would indicate a greater \textit{richness} of reservoir activation curves to draw a set of predictions from. Between the MF and non-MF ERRC heat maps, we observe a slightly larger population of neurons with ``high'' (above 60) unique local maxima; however, the differences are less pronounced. Between the ERRC and FFRC models, we see that more neurons are involved in the ERRC predictions overall in both MF and non-MF cases. Moreover, in the ERRC there are neurons with a higher magnitude of unique local maxima (roughly 70 in the ERRC versus 45 in the FFRC). This follows intuitively: a small subset of neurons in the fly brain act as ``information highways for multisensory integration'' \cite{Mehta2022-vf}; conversely, randomly-weighted neurons have arbitrary importance, and thus we would expect that an effective output prediction would rely on a broad sampling of activations in order to match to a ground truth signal. Finally, across all figures in Fig.\,\ref{fig:maxcounts}, we note that increasing the spectral radius $\rho$ also increases the proportion of neurons which have a high number of unique local maxima.

\subsubsection{Experiment 4}\label{sssec:Exp4results_bifplots}
\begin{figure*}[h]
\begin{subfigure}[h]{0.47\linewidth}
\centering
\includegraphics[width=0.9\linewidth]{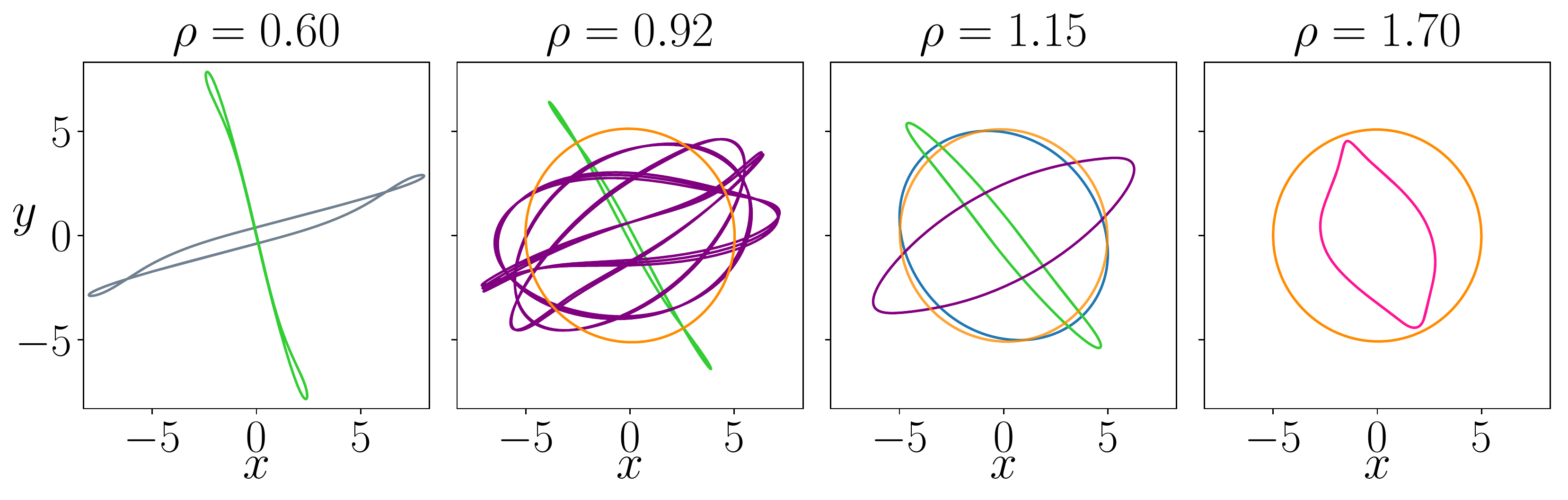}
\includegraphics[width=0.99\linewidth]{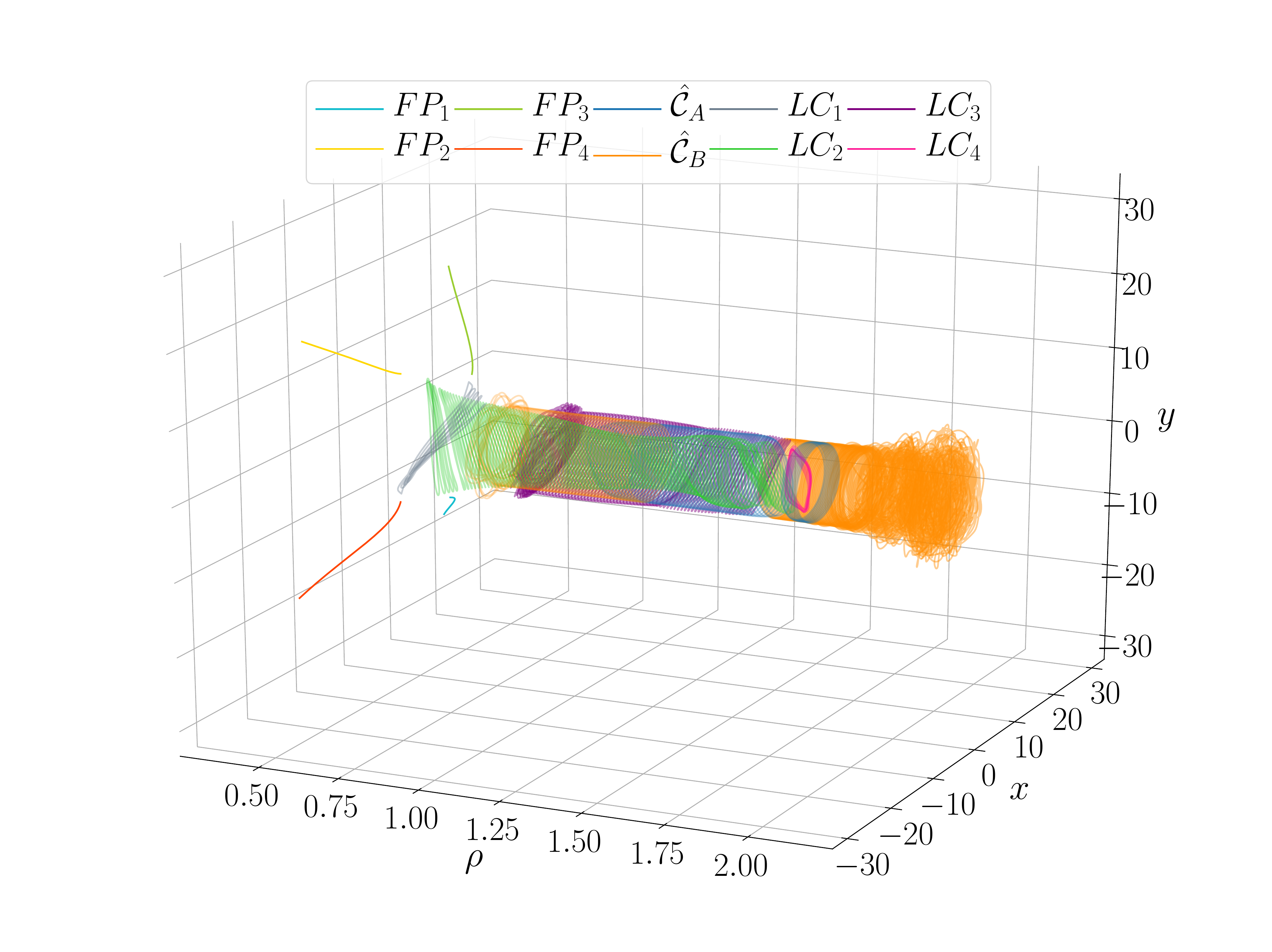}
\caption{ERRC}
\label{fig:randdynamics}
\end{subfigure}
\hfill
\begin{subfigure}[h]{0.47\linewidth}
\centering
\includegraphics[width=0.9\linewidth]{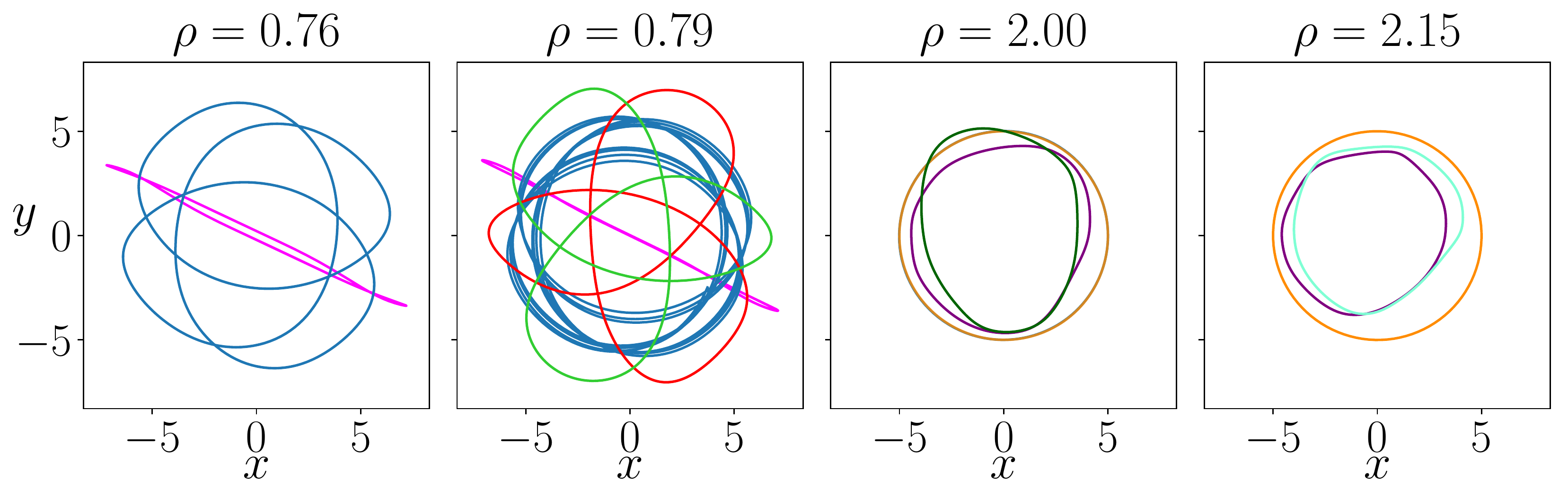}
\includegraphics[width=0.99\linewidth]{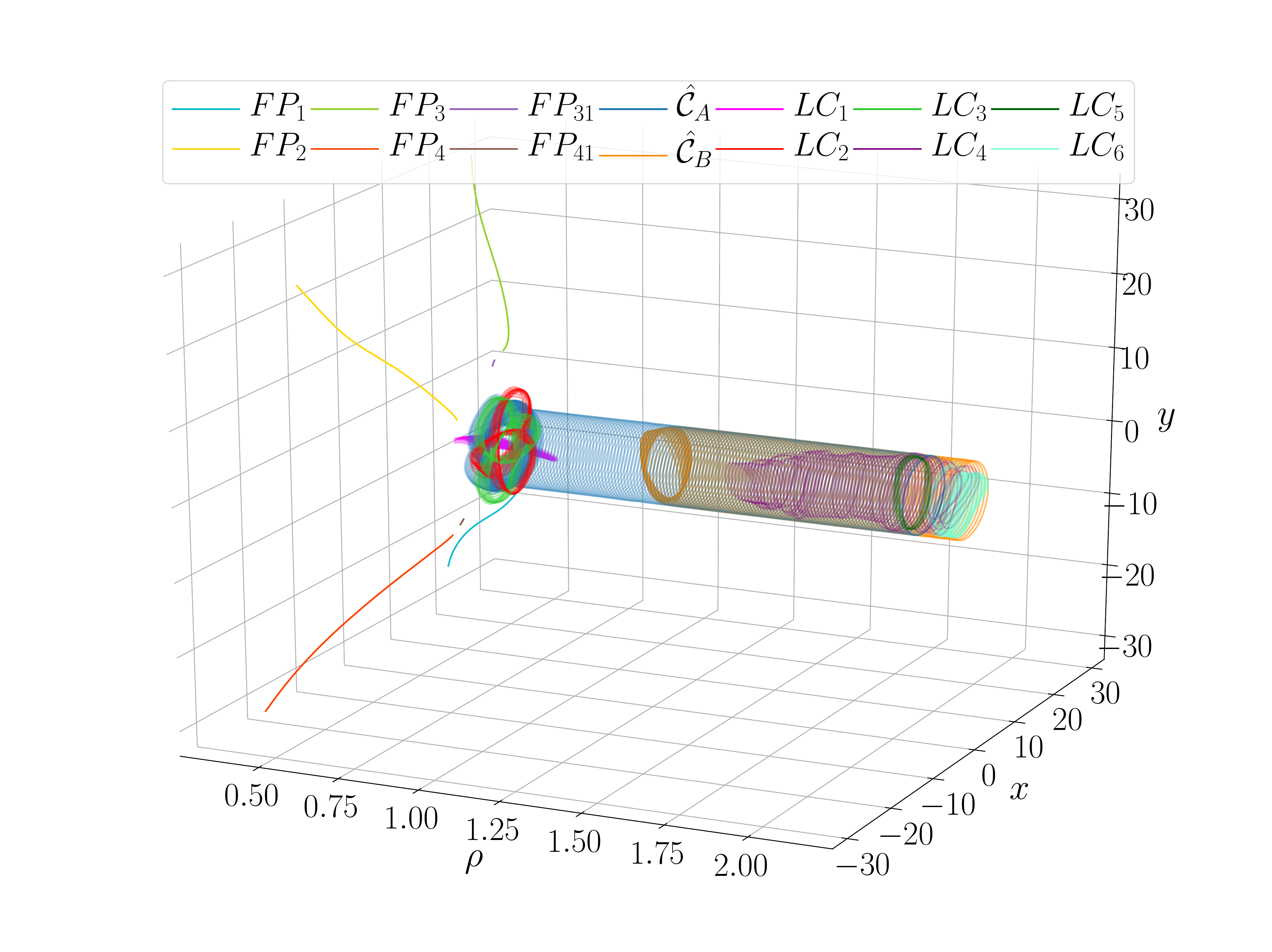} 
\caption{FFRC}
\label{fig:ffrcdynamics}
\end{subfigure}
\caption{Dynamics of the ERRC in (a) and FFRC in (b) for increasing $\rho$. $FP_i$: fixed point $i$; $LC_j$: limit cycle $j$; $\hat{\mathcal{C}}_{A}$: reconstructed $\mathcal{C}_{A}$; $\hat{\mathcal{C}}_{B}$: reconstructed $\mathcal{C}_{B}$.}
\label{fig:bifplots}
\end{figure*}

We now explore the prediction dynamics of the ERRC and FFRC in the respective $\mathbb{P}$ for increasing $\rho$. 

For the ERRC, (see Fig. \ref{fig:randdynamics}), like in \cite{flynn2023seeingdouble}, at small $\rho$ values we find that four anti-symmetric fixed points exist which subsequently bifurcate into two distinct limit cycles ($LC$) at $\rho \approx 0.55$, whose dynamics are shown above the bifurcation diagram for $\rho=0.6$. $LC_{1}$ can no longer be tracked for $\rho > 0.62$, while $LC_2$ remains stable and exists up to $\rho=1.65$. Coexisting with $LC_2$ here is the reconstruction of $\mathcal{C}_{B}$, which first appears as a torus at $\rho=0.76$ and begins to more closely resemble $\mathcal{C}_{B}$ from $\rho=0.78$ to $\rho=1.25$. The state of the ERRC subsequently tends to $LC_{3}$, which is born at $\rho=0.91$; this is initially a torus before it becomes a limit cycle -- as highlighted by the plots in $\mathbb{P}$ at $\rho=0.92,1.15$ (above the bifurcation diagram in Fig.\,\ref{fig:randdynamics}). $LC_{3}$ can no longer be tracked for $\rho>1.69$. At $\rho=1.64$, $\hat{\mathcal{C}}_{B}$ is reborn; however, for $\rho>1.83$ and beyond, $\hat{\mathcal{C}}_{B}$ becomes chaotic, where it remains indefinitely (or for as far as it can be tracked with reasonable accuracy). 

The ERRC is shown to exhibit \textit{multifunctionality} -- wherein $\hat{\mathcal{C}}_{A}$ is found to coexist with $\hat{\mathcal{C}}_{B}$ -- for the intervals of $\rho=[1.17,1.25]$ and $[1.71,1.76]$. An additional limit cycle, $LC_{4}$, also briefly appears here, for $\rho=[1.69,1.71]$, and its dynamics for $\rho=1.7$ is shown above the bifurcation diagram.

In Fig.\,\ref{fig:ffrcdynamics} we illustrate the prediction dynamics of the FFRC on the seeing double problem. We find that, as in Fig.\,\ref{fig:randdynamics}, at small $\rho$ values, four anti-symmetric fixed points exist. However, as we continue to track the changes in these fixed points the differences between how the FF and ER RC solves the seeing double problem emerges. We find that for $FP_{3}$ and $FP_{4}$ there is a small region of hysteresis with an additional branch of stable fixed points -- $FP_{31}$ and $FP_{41}$, respectively. For $\rho>0.7$, $FP_{31}$ and $FP_{41}$ can no longer be tracked and the state of the RC tends to $FP_{1}$ and $FP_{2}$, respectively. At $\rho=0.73$ there is a bifurcation from these fixed points to a limit cycle, $LC_{1}$ (potentially a SNIPER/homoclinic bifurcation based on the characteristics of $LC_{1}$). For $\rho>0.84$, $LC_{1}$ can no longer be tracked and the state of the FFRC tends to $\hat{\mathcal{C}}_{A}$, which exists up to $\rho=2.08$. By tracking $\hat{\mathcal{C}}_{A}$ as $\rho$ decreases, we find that it goes through several bouts of torii bifurcations, initially appearing as a period-3 limit cycle at $\rho=0.75$, highlighted by the plot above the bifurcation diagram in Fig.\,\ref{fig:ffrcdynamics} for $\rho=0.76$, we also show here for $\rho=0.79$ that there exists two antisymmetric limit cycles $LC_{2}$ and $LC_{3}$ for $\rho=[0.77,0.80]$.

We see in Fig.\,\ref{fig:ffrcdynamics} that the FFRC achieves \textit{multifunctionality} for $\rho=[1.29,2.08]$; even though $\mathcal{C}_{B}$ comes into existence at $\rho=1.27$, it is only properly reconstructed (according to the criteria in Sec.\,\ref{ssec:seeingdouble}) at $\rho = 1.29$, and coexists with $\mathcal{C}_{A}$ until $\rho=2.08$. We also note that another limit cycle ($LC_{4}$) exists while the FFRC is multifunctional. It is suggested that additional limit cycles exist in $\mathbb{P}$ as $\rho$ continues to increase.

Comparing the findings for the ERRC and FFRC, what is perhaps most interesting is that we observe clear evidence of multifunctionality for a much broader range of row values -- from $\rho=1.29$ to $2.08$, totalling a multifunctional $\rho$-interval of $~0.79$, which is much larger than that of the ERRC ($\approx0.13$). Moreover, compared to the ERRC -- which becomes chaotic at large $\rho$ values -- the FFRC prediction dynamics persist (without succumbing to chaos) long past this $\rho$ limit.

\section{Conclusion}\label{sec:conclusion}
\subsection{Summary}
In this paper we find that the FFRC outperforms the ERRC on the seeing double problem  in \textbf{three ways}: 
\begin{enumerate}
    \item A higher frequency of achieving multifunctionality for $\rho=1.4$, $\gamma=5$ (Experiment 1).
    \item A broader window of $[\rho,\gamma]$-space where multifunctionality is present, and a larger magnitude of multifunctionality overall (Experiment 2).
    \item Prediction dynamics of the FFRC persist as non-chaotic, circular trajectories well beyond the observed $\rho$ threshold in the ERRC (Experiment 4). 
\end{enumerate}

These findings suggest that fruit fly brain structure -- relative to an arbitrary, random topology -- possesses a \textit{greater capacity for multifunctionality}, and is \textit{more robust} to MF-related parameter fluctuations. Interestingly, the FFRC appears to mimic analogous abilities observed in its biological counterpart \cite{OSullivan2018-ky}. 

Fig.\,\ref{fig:maxcounts} suggests that the FFRC takes advantage of a smaller population of neurons with higher importance -- i.e. for a given $\rho$ value there are only a small number of neurons which `fire' with a large number of unique local maxima. In comparison, we see here that for the ERRC there is a larger proportion of highly activated neurons for a given $\rho$.

\subsection{Limitations}
We acknowledge that our FFRC adjacency matrix, $\textbf{M}_{FF}$, is a \textit{translation} of its corresponding connectome ROI; which captures broad lateral horn connectivity, but does not include all synapses -- i.e. due to applying a threshold (see Sec.\,\ref{sec:Methods}). As a structural translation only, the FFRC also fails to capture \textit{functional} aspects (e.g. the action of neurotransmitters). Detailed structural elements are also absent, such as neuron morphologies (we use point neurons). One could therefore argue that this structurally-inspired model is only providing a ``taste'' of the potential capacity of the fly brain for multifunctionality.

\subsection{Future Work}
We will continue to analyse the nonlinear and chaotic dynamics of the FFRC and ERRC in order to further explore their limits of multifunctionality. We will also seek to (as in \cite{flynn2022_MFLimits}) determine the impact of additional model factors -- such as the input matrix designs ($\mathbf{W}_{in}$) -- on multifunctionality.

We aim to transplant other animal connectome-based networks, such as \cite{Cook2019-nb,Oh2014-vz}, to RC setups to test whether these networks can also be exploited in a machine learning context.
\vspace{.6cm}
\bibliography{bib}
\end{document}